\documentclass[runningheads]{llncs}
\usepackage{graphicx}

\usepackage{tikz}
\usepackage{comment}
\usepackage{amsmath,amssymb} 
\usepackage{tabularray-2021}
\usepackage{hyperref}

\usepackage[accsupp]{axessibility}  

\begin{document}
\pagestyle{headings}
\mainmatter
\def\ECCVSubNumber{3}  

\title{CCRL: Contrastive Cell Representation Learning} 

\titlerunning{CCRL: Contrastive Cell Representation Learning}
%
\author{Ramin Nakhli\inst{1} \and
Amirali Darbandsari\inst{1} \and
Hossein Farahani\inst{1} \and
Ali Bashashati\inst{1}}
\authorrunning{R. Nakhli et al.}
%
\institute{University of British Columbia \\
\email{\{ramin.nakhli,a.darbandsari, h.farahani,ali.bashashati\}@ubc.ca}}
\maketitle

\begin{abstract}
Cell identification within the H\&E slides is an essential prerequisite that can pave the way towards further pathology analyses including tissue classification, cancer grading, and phenotype prediction. However, performing such a task using deep learning techniques requires a large cell-level annotated dataset. Although previous studies have investigated the performance of contrastive self-supervised methods in tissue classification, the utility of this class of algorithms in cell identification and clustering is still unknown. In this work, we investigated the utility of Self-Supervised Learning (SSL) in cell clustering by proposing the Contrastive Cell Representation Learning (CCRL) model. Through comprehensive comparisons, we show that this model can outperform all currently available cell clustering models by a large margin across two datasets from different tissue types. More interestingly, the results show that our proposed model worked well with a few number of cell categories while the utility of SSL models has been mainly shown in the context of natural image datasets with large numbers of classes (e.g., ImageNet). The unsupervised representation learning approach proposed in this research eliminates the time-consuming step of data annotation in cell classification tasks, which enables us to train our model on a much larger dataset compared to previous methods. Therefore, considering the promising outcome, this approach can open a new avenue to automatic cell representation learning.

\end{abstract}

\section{Introduction}

Cells are the main components that determine the characteristics of tissues and, through mutual interactions, can play an important role in many aspects including tumor progression and response to therapy~\cite{heindl2015mapping,son2017role,levy2020spatial}. Therefore, cell identification can be considered as the first and essential building block for many tasks, including but not limited to tissue identification, slide classification, T-cell infiltrating lymphocytes analysis, cancer grade prediction, and clinical phenotype prediction~\cite{javed2020cellular,martin2021predictive}. In clinical practice, manual examination of the Whole Slide Images (WSI), multi-gigapixel microscopic scans of tissues stained with Hematoxylin \& Eosin (H\&E), is the standard and widely available approach for cell type identification~\cite{alturkistani2016histological}. However, due to the large number of cells and their variability in texture, not only is the manual examination time-consuming and expensive, but it also introduces intra-observer variability~\cite{boyle2000abc}. Although techniques such as immunohistochemistry (IHC) staining can be used to identify various cell types in a tissue, they are expensive, are not routinely performed for clinical samples, and require a deep biological knowledge for biomarker selection~\cite{van1986cell}.  \par

With the exponential growth of machine learning techniques in recent years, especially deep learning, computational models have been proposed to accelerate the cell identification process~\cite{sirinukunwattana2016locality,graham2019hover}. However, these models need large cell-level annotated datasets to be trained on~\cite{graham2019hover,amgad2021nucls}. Collecting such datasets is time-consuming and costly as the pathologists have to annotate tens of thousands of cells present in H\&E slides, and this procedure has to be carried out for any new tissue type. To mitigate this, several research groups have moved to crowd-sourcing~\cite{amgad2021nucls}. Nevertheless, providing such pipelines is still difficult in practice. Therefore, all the aforementioned problems set a strong barrier to the important cell identification task which is a gateway to more complex analyses. \par

In this paper, using state-of-the-art deep learning methods, we investigate the utility of contrastive self-supervised learning to obtain representations of cells without any kind of supervision. Moreover, we show that clusters of these representations are associated with specific types of cells enabling us to apply the proposed model for cell identification on routine H\&E slides in a massive scale. Furthermore, our results demonstrate that our trained model can outperform all the existing counterparts. \par

Therefore, the contributions of this work can be summarized as: 1) the first work to study the utility of self-supervised learning in cell representation learning in H\&E images; 2) introducing a novel framework for this purpose; 3) outperforming all the existing unsupervised baselines with a large margin. The code will be available at \href{https://github.com/raminnakhli/Contrastive-Cell-Representation-Learning}{https://github.com/raminnakhli/Contrastive-Cell-Representation-Learning}. \par

\section{Previous Works}

\subsection{Self-Supervised Learning} 

Self-Supervised Learning (SSL) is a technique to train a model without any human supervision in a way that the generated representations capture the semantics of the image. Learning from the pseudo-labels generated by applying different types of transformations has been a popular approach in the early ages of this technique. Learning local position of image patches~\cite{noroozi2016unsupervised}, rotation angle prediction~\cite{komodakis2018unsupervised}, and color channel prediction~\cite{zhang2016colorful} are some of the common transformations used to this end. \par

Recent studies have shown that using similar transformations in a contrastive setting can significantly enhance the quality of the representations to an extent that makes it possible to even outperform supervised methods after fine-tuning~\cite{chen2020improved}. In this setting, models pull the embeddings of two augmentations of the same image (one called query and the other called positive sample) together while they push the embedding of other images (negative samples) as far as possible. SimCLR~\cite{chen2020simple,chen2020big} proposes using the same encoder network to encode the query, positive, and negative samples while MoCo~\cite{he2020momentum,chen2020improved} introduces using the momentum encoder to encode the positive and negative samples, which is updated by the weights of the query image encoder. On the other hand, BYOL~\cite{grill2020bootstrap} and DINO~\cite{caron2021emerging} remove the need for negative samples. \par

In addition to the applications of the self-supervised learning in image classification~\cite{chen2020big,chen2020improved,caron2021emerging} and object detection~\cite{xie2021detco}, several papers have shown benefits of this approach in medical imaging. Azizi et al.~\cite{azizi2021big} show their model can outperform supervised models across multiple medical imaging datasets by pulling the embeddings of two different views from the same patient under different conditions, and~\cite{xie2020instance} applies self-supervised learning for cell detection. while~\cite{xie2020instance} applies self-supervised learning for cell segmentation,~\cite{ciga2022self} uses SimCLR to learn representations on patches of H\&E slides. Furthermore, they show that increasing the dataset size and variety improves the performance of the model on the patch classification task. In contrast,~\cite{zhang2021histopathology} shows that using another type of contrastive learning architecture reduces the final performance of patch classification compared to the pre-trained models on the ImageNet dataset~\cite{deng2009imagenet}. Although some of the aforementioned studies investigate the influence of contrastive self-supervised learning in the context of histopathology patch classification, to the best of our knowledge, the applications of self-supervised techniques for cell labelling (as opposed to patch classification) are largely ignored. Furthermore, the contradictory results of the aforementioned studies (one showing the superiority of SSL over the ImageNet pre-trained model, and the other showing the opposite) warrant further investigation of these models in the context of cell clustering and labelling in histopathology. \par

\subsection{Cell Classification in Histopathology}

The utility of machine learning algorithms in cell classification is an active and important area of research. Earlier works have been mainly focused on extracting hand-crafted features from cell images and applying machine learning classification models to perform this task. For example, using the H\&E images,~\cite{nguyen2011prostate} extracts cytological features from cells and applies Support Vector Machines (SVM) to separate the cell types. ~\cite{dalle2009nuclear} uses the size, color, and texture of the cells to assign a score to each cell based on which they can classify the cell. Later works combined the deep and hand-crafted features to improve the accuracy of cell classification~\cite{cruz2013deep}. However, recent studies (for example,~\cite{sirinukunwattana2016locality,graham2019hover}) show that cell classification accuracy can significantly improve, solely based on deep learning-based features. \par

Although these studies have shown promising performance, they require a large annotated dataset which is difficult to collect. A few recent studies have focused on unsupervised cell classification to address this problem. For example, Hue et al.~\cite{hu2018unsupervised} take advantage of the InfoGAN~\cite{chen2016infogan} design to provide a categorical embedding for images, based on which they can differentiate between cell types and Vununu et al.~\cite{vununu2020strictly} propose using Deep Convolution Auto-encoder (DCAE) which learns feature embeddings by performing image reconstruction and clustering at the same time. However, all of these works are focused on only one tissue type. In this study, we present the first contrastive self-supervised cell representation learning framework using H\&E images and show that this design can consistently outperform all the currently available baselines across two tissue types. \par

\section{Method}

\begin{figure}
\includegraphics[width=\textwidth]{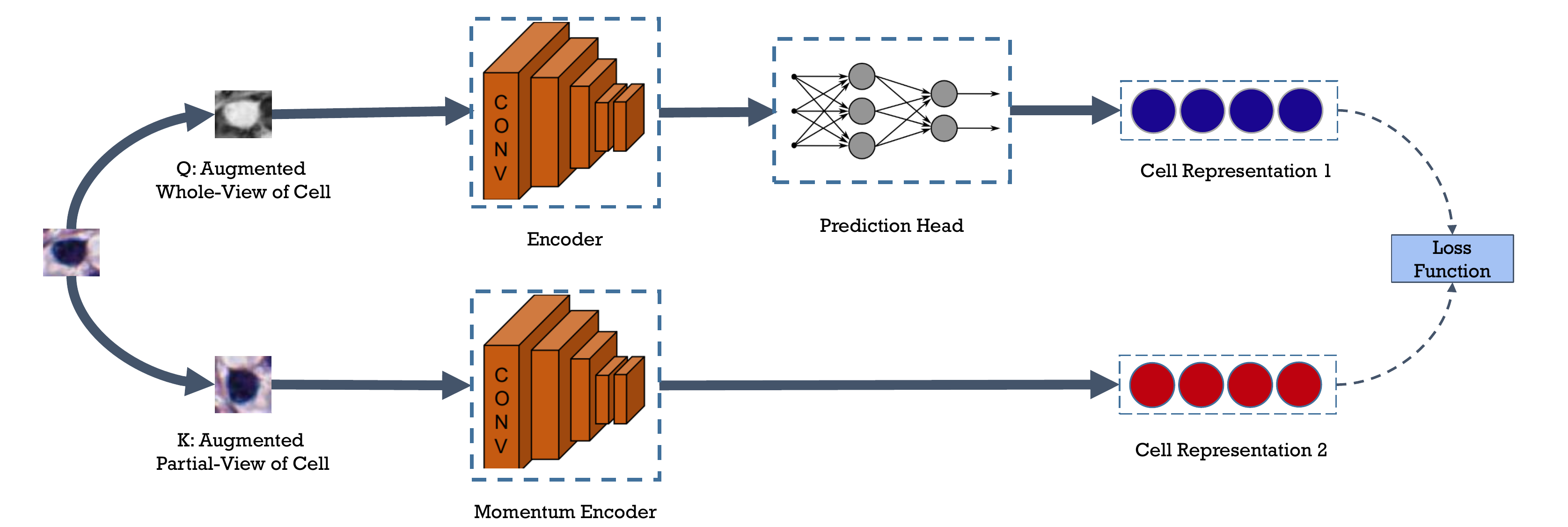}
\caption{Overview of the Framework} \label{fig1}
\end{figure}

Given an image of a cell, our objective is to learn a robust representation for the cell which can be used for down-stream tasks such as cell clustering. Fig.~\ref{fig1} depicts an overview of our proposed self-supervised method, called CCRL (Contrastive Cell Representation Learning). The main goal of our framework is to provide the same representation for different views of a cell. More specifically, this framework consists of two branches, query and key, which work on different augmentations of the same image. \par

\noindent In this design, cell embeddings are learned by pulling the embeddings of two augmentations of the same image together, while the representations of other images are pushed away. Consider the input image batch of $X=x_1, x_2, ..., x_N$ where $x_i$ is a small crop of the H\&E image around a cell in a way that it only includes that specific cell. Two different sets of augmentations are applied to $X$ to generate $Q=\{q_i|i = 1,...,N\}$ and $K=\{k_i|i = 1,...,N\}$ where $N$ is the batch size. These sets are called query and key, respectively, and $q_i$ and $k_j$ are the augmentations of the same image if and only if $i=j$. The query batch is encoded using a backbone model, a neural network of choice, while the keys are encoded using a momentum encoder, which has the same architecture as the backbone. Using a momentum encoder can be viewed as keeping an ensembling of the query model throughout its training, providing more robust representations. This momentum encoder is updated using the equation~\ref{moemtum_eq} in which $\theta_{k}^{t}$ is the parameter of momentum encoder at time $t$, $m$ is the momentum factor, and $\theta_{q}^{t}$ is the parameter of the backbone at time $t$

\begin{equation} \label{moemtum_eq}
    \theta_{k}^{t} = m\theta_{k}^{t-1} + (1-m)\theta_{q}^{t}
.\end{equation}

\noindent Consequently, the obtained query and key representations are passed through separate Multi-Layer Perceptron (MLP) layers called projector heads. Although the query projector is trainable, the key projector is updated with momentum using the weight of the query projector head. Similar to the momentum encoder, the key projector can be considered as an ensembling of that of the query branch. We have restricted these models to be 2-layer MLPs with the input size of 512, hidden size of 128, and output size of 64. In addition to the projector head, we use an extra MLP on the query side of the framework, called prediction head. This extra network enables us to provide asymmetricity in the design of our model (as apposed to using a deeper projector head which keeps the design symmetric), providing more flexible representations on the query branch for competing with the ensembled representations coming from the key branch. This network is a 2-layer MLP with input, hidden, and output sizes of 64, 32, and 64, respectively. Similar to the last fully-connected layers of a conventional classification network, the projection and prediction heads provide more representation power to the model. \par

Finally, the models are trained using the equation~\ref{infonce_eq}, pulling the positive and pushing the negative embeddings

\begin{equation} \label{infonce_eq}
    L_{q_{i}}^{cell} = -\log \frac{\exp\frac{\Vert f_q(q_i) \Vert^2 . \Vert f_k(k_i) \Vert^2}{\tau}}{\sum_{j=0}^{N+Q} \exp\frac{\Vert f_{q}(q_i) \Vert^2 . \Vert f_k(k_j) \Vert^2}{\tau}}
.\end{equation}

\noindent In this equation, $\tau$ is the temperature which controls the sharpness of the similarity distribution, $Q$ is the number of items stored in the queue from the key branch, $\Vert x \Vert^2$ is the second-order normalization of $x$, $f_q$ is the equal function for the combination of the backbone, query projection head, and query prediction head, and $f_k$ shows the equal function for the momentum encoder and the key projection head. \par

As mentioned above, we use an external memory bank to store the processed key representations. The stored representation will be used in the contrastive setting as negative samples, and they are limited to $65,536$ samples throughout the training.

Additionally, we incorporated a local-global connection technique to ensure that the model is always focusing on the whole view representation of the cell throughout the training process. To this end, only one of the two augmentation pipelines includes cropping operations. This pipeline generates local regions of the cell image while the images generated by the other augmentation pipeline are global, containing the whole-cell view. The rest of the operations are the same in both pipelines, and they include color jitter (brightness of 0.4, contrast of 0.4, saturation of 0.4, and hue of 0.1), gray-scale conversion, Gaussian blur (with a random sigma between 0.1 and 2.0), horizontal and vertical flip, and rotation (randomly selected between 0 to 180 degrees). \par

At inference time, cell embeddings were generated from the trained momentum encoder and were clustered by applying the K-means algorithm. It is worth mentioning that one can use either the encoder or momentum encoder for embedding generation; however, the momentum encoder provides more robust representations since it aggregates the learned weights of the encoder network from all of the training steps (an ensembling version of the encoder throughout training). We refer to this technique as ensembling in the rest of this article. \par

\section{Experiments}

\subsection{Evaluation Metrics}

As the main goal of this work was to provide a framework for clustering of cell types, we evaluated the performance of the models using Adjusted Mutual Information (AMI), Adjusted Random Index (ARI), and Purity of the identified cell clusters by the model and the ground truth labels. 

\subsubsection{AMI} captures the agreement between two sets of assignments using the amount of the mutual information that exists between these sets. However, it is adjusted to mitigate the effect of chance in the score. \par

\subsubsection{ARI} is the chance adjusted form of the Rand Index, which calculates the quality of the clustering based on the number of instance pairs. \par

\subsubsection{Purity} measures how the samples within each cluster are similar to each other. In other words, it demonstrates if each cluster is a mixture of different classes. \par

\subsection{Datasets}

To demonstrate the utility and performance of our proposed model, we used two publicly available datasets  (CoNSeP~\cite{graham2019hover} and NuCLS~\cite{amgad2021nucls}) representing two different tissue types with cell-level annotations. Although the annotations were not used in the training step, we leveraged them to evaluate the performance of different models on the test set. \par

The CoNSeP dataset consisted of 41 H\&E tiles from colorectal tissues extracted from 16 whole slide images of a single patient. All tiles were in 40$\times$ magnification scale with the size of 1,000$\times$1,000 pixels. Cell types included 7 different categories of normal epithelial, malignant epithelial, inflammatory, endothelial, muscle, fibroblast, and miscellaneous. However, as suggested by the original paper, normal and malignant epithelial were grouped into the epithelial category, and the muscle, fibroblast, and endothelial cells were grouped into the spindle-shaped category. Therefore, the final 4 groups included epithelial, oval-shaped, inflammatory, and miscellaneous. \par

The NuCLS dataset included 1744 H\&E tiles of breast cancer images from the TCGA dataset collected from 18 institutions. The tiles had different sizes, but they were roughly 300$\times$300 pixels. There were 12 different cell types available in this dataset: tumor, fibroblast, lymphocyte, plasma, macrophage, mitotic, vascular endothelium, myoepithelium, apoptotic body, neutrophil, ductal epithelium, and eosinophil. However, as suggested by the paper, these subtypes were grouped together into 5 superclasses including tumor (containing tumor and mitotic cells), stromal (containing fibroblast, vascular endothelium, and macrophage), sTILs (containing lymphocyte and plasma cells), apoptotic cells, and others. \par


Details of each dataset can be found in Table~\ref{dataset_tb}. \par 

\begin{table}
\caption{Dataset details}
\label{dataset_tb}
\centering
\begin{tblr}{ Q[c,m,2cm] Q[c,m,2cm] Q[c,m,2cm]}
\hline
 & CoNSeP & NuCLS \\
\hline
Type Count &  4 & 5\\
Cell Count & 24,319 & 51,986\\
Tile Count & 41 & 1,744\\
Tile Size & 1,000$\times$1,000 & 300$\times$300\\
Tissue Type & Colorectal & Breast\\
\hline
\end{tblr}
\end{table}

\subsection{Data Preparation}

The aforementioned datasets included patch-level images, while we required cell-level ones for the training of the model. To generate such data, we used the instance segmentation provided in each of the datasets to find cells and crop a small box around them. We adopted an adaptive window size to extract these images, whose size was equal to twice the size of the cell in CoNSeP and equal to the size of the cell in the NuCLS dataset. The images were resized to $32\times32$ pixels before being fed into our proposed framework. \par


\subsection{Implementation Details}

The code was implemented in Pytorch, and the model was run on a V100 GPU. The batch size was set to 1024, the queue size to $65536$, and pre-activated ResNet18~\cite{he2016identity} was used for the backbone. The model was trained using Adam optimizer for 500 epochs with a starting learning rate of 0.001, a cosine learning rate scheduler, and a weight decay of 0.0001. We also adopted a 10-epoch warm-up step. The momentum factor in the momentum encoders was set to 0.999, and the temperature was set to 0.07. \par

\subsection{Results}

The results of unsupervised clustering of CCRL can be found in Table~\ref{results_tb} as well as that of the baselines. We compared the performance of our model with five different baseline and state-of-the-art models. The pre-trained ImageNet used the weights trained on the ImageNet dataset to generate the cell embeddings. The second baseline model used morphological features to produce a 30-dimensional feature vector, consisting of geometrical and shape attributes~\cite{bhaskar2019methodology}. The third baseline method utilized the Manual Features (MF), a combination of Scale-Invariant Feature Transform (SIFT) and Local Binary Patterns (LBP) features, proposed by~\cite{hu2018unsupervised}. We also compared the results of our model with two state-of-the-art unsupervised cell clustering methods. The DCAE~\cite{vununu2020strictly} model adopted a deep convolution auto-encoder model alongside a clustering layer to learn cell embeddings by preforming an image reconstruction task. Also, the authors of~\cite{hu2018unsupervised} developed a generative adversarial model for cell clustering by increasing the mutual information between the cell representation and a categorical noise vector. \par

\begin{table}
\caption{Unsupervised clustering performance comparison.}
\label{results_tb}
\centering
\resizebox{\textwidth}{!}{

\begin{tblr}{ Q[c, m, 3cm] Q[c, m, 1cm] Q[c, m, 1cm] Q[c, m, 1cm] Q[c, m, 1cm] Q[c, m, 1cm] Q[c, m, 1cm]}
\hline
 \multirow{2}{*}{Model} & \multicolumn{3}{|c|}{CoNSeP} & \multicolumn{3}{|c|}{NuCLS}\\
\hline
 & AMI & ARI & Purity & AMI & ARI & Purity\\
\hline
Pre-trained ImageNet & 7.3\% & 7\% & 42.7\% & 9.3\% &7.8\% & 56.7\%\\
Morphological~\cite{bhaskar2019methodology} & 12.7\% & 1.3\% & 48.8\% & 21.1\% & 18.8\% & 66.1\%\\
Manual Features~\cite{hu2018unsupervised} & 9.5\% & 6.4\% & 45.5\% & 11.25\% & 7.8\% & 56.2\%\\
Auto-Encoder~\cite{vununu2020strictly} & 10.1\% & 7.3\% & 50.5\% & 8.3\% & 7.2\% & 56.8\%\\
InfoGAN~\cite{hu2018unsupervised} & 14.8\% & 15.7\% & {\bfseries 58.4\%} & 14\% & 12.6\% & 62\%\\
CCRL (Ours) & {\bfseries 24.2\%} & {\bfseries 21.7\%} & 51.8\% & {\bfseries 22.8\%} & {\bfseries 24\%} & {\bfseries 68.3\%}\\
\hline
\end{tblr}}
\end{table}

As can be seen in Table~\ref{results_tb}, our model can outperform its counterparts with a large margin in terms of different clustering metrics across all datasets. \par

\subsection{Ablation Study}

Ablation studies were performed on three most important components of our framework: 1) local-global connection technique; 2) inference with the ensemble model; 3) query prediction head. \par

Tables~\ref{ablation_tb} demonstrates the effect of ablation of each component. Based on these experiments, all components are essential for learning effective cell representations regardless of the tissue type. In this part, we used NuCLS dataset for ablation, as it has been collected from multiple patients and diverse locations. \par

\begin{table}
\caption{Ablation study. First and second performances are highlighted and underlined, respectively.}
\label{ablation_tb}
\centering
\begin{tblr}{ Q[c,m] Q[c,m] Q[c,m] Q[c,m] Q[c,m] Q[c,m] Q[c,m]}
\hline
                    & AMI                   & ARI               & Purity            \\
\hline
w/o Local-Global    & 21.9\%                & 20.4\%            & \textbf{69.7\%}   \\
w/o Ensembling      & 21.1\%                & 19.2\%            & 67.2\%            \\
w/o Prediction Head & \underline{22\%}      & \underline{20.9\%}& 68.1\%            \\
w/ All              & \textbf{22.8\%}       & \textbf{24\%}     & \underline{68.3\%}\\
\hline
\end{tblr}
\end{table}

\section{Discussion \& Conclusion}

Cell identification is a gateway to many complex tissue analysis applications. However, due to the large number of cells in H\&E slides, manual execution of such a task is very time-consuming and resource-intensive. Although several research studies have provided machine learning models to classify cells in an automatic manner, they still require a large dataset that is manually annotated. In this paper, we investigated the utility of self-supervised learning in the context of cell representation learning by designing a self-supervised model with designated architecture for the task of cell representation learning. The quality of the representations was measured based on the clustering performance, by applying the K-means algorithm on top of these representations and measuring the cluster enrichment in specific cell types. Our experiments confirm that the SSL training improves the clustering metrics compared to currently available unsupervised methods. \par

It is worthwhile to mention that SSL frameworks are mostly evaluated on natural images (e.g., ImageNet dataset), which include a large number of categories. However, in our case, the number of classes (i.e., cell types) is extremely small (4 and 5 classes for the CoNSeP and NuCLS datasets, respectively, versus 1000 classes for ImageNet). This means that, on average, 25\% and 20\% of the negative samples for the CoNSeP and NuCLS dataset are false-negatives, respectively, while this ratio is only 0.1\% for the ImageNet dataset. Therefore, our findings show that the proposed SSL framework can operate well when a small number of classes or categories exists. \par

This paper is the first attempt to apply contrastive self-supervised learning to cell identification in H\&E images. The proposed model enables us to achieve robust cell representation using an enormous amount of unlabeled data which can simply be generated by scanning routine H\&E stained slides in the clinical setting. Furthermore, due to the unsupervised learning nature of the framework, the proposed model has the potential to identify novel cell types that may have been overlooked by pathologists. In addition to the above-mentioned benefits that an SSL framework could provide in the context of cell classification, these models are also robust to long-tail distributions in the data~\cite{liu2021self}; hence addressing the common issue of rare cell populations in pathology (e.g., tumor budding, mitotic figures). Therefore, we hope that this work motivates researchers and serves as a step towards more utilization of unsupervised learning in pathology applications, especially in the context of cell-level information representation. \par

%
%



\end{document}